\documentclass[times, review, 10pt]{elsarticle}

\usepackage{amssymb}
\usepackage{amsmath}
\usepackage{cases}
\usepackage{booktabs}
\usepackage{makecell}
\usepackage{tabularx}
\usepackage{multirow}
\usepackage{array}

\usepackage{textcomp}
\usepackage{stfloats}
\usepackage{url}
\usepackage{graphicx}
\usepackage{booktabs}
\usepackage{xcolor}
\usepackage{colortbl}
\usepackage{changepage}
\usepackage{algorithm}
\usepackage{algpseudocode}
\usepackage{soul}

\usepackage{comment}

\usepackage{subcaption}

\journal{Computer Vision and Image Understanding}

\begin{document}

\begin{frontmatter}



\title{SAGA: Selective Adaptive Gating for Efficient and Expressive Linear Attention}


\author[t1,t2]{Yuan Cao}
\ead{23115050@bjtu.edu.cn}
\author[t1,t2]{Dong Wang \corref{cor1}}
\ead{wangdong@bjtu.edu.cn}

\affiliation[t1]{organization={Institute of Information Science},
            addressline={Beijing Jiaotong University}, 
            city={Beijing},
            postcode={100044}, 
            country={China}}

\affiliation[t2]{organization={Beijing Key Laboratory of Advanced Information Science and Network Technology},
            city={Beijing},
            postcode={100044}, 
            country={China}}
\cortext[cor1]{Corresponding author}

%% Abstract
\begin{abstract}
Recent advances in Vision Transformers demonstrate strong long-range modeling ability, but the quadratic complexity of softmax-based attention severely limits their scalability to high-resolution vision tasks. Linear attention offers a promising alternative by reformulating the attention computation from $(QK)V$ to $Q(KV)$, reducing complexity from $\mathcal{O}(N^2)$ to $\mathcal{O}(N)$ while maintaining global context. However, most linear attention methods uniformly compress key-value (KV) representations, producing low-rank feature maps that constrain expressivity. To address this, we propose \textbf{SAGA}, which adaptively introduces a gating matrix matching the dimensions of the KV feature maps to enable fine-grained modulation of token contributions. This mechanism enhances semantic diversity, mitigates redundancy and effectively increases matrix rank. Furthermore, we design an efficient Hadamard-product decomposition for gate computation, introducing negligible computational or memory overhead. Extensive experiments show that SAGA improves ImageNet-1K Top-1 accuracy by \textbf{+1.1\%}  over MLLA and consistently enhances downstream detection and segmentation. For low-light enhancement, SAGA reduces runtime and GPU memory by over \textbf{80\%} compared with LLFormer, with negligible degradation in quality. Overall, SAGA provides an expressive yet lightweight linear-attention formulation for scalable vision models.

\end{abstract}

%% Keywords

\end{frontmatter}

\section{Introduction}
In recent years, the Vision Transformer (ViT)~\cite{dosovitskiy2020image} has emerged as a leading architecture for visual recognition, largely due to its strong ability to model long-range dependencies. By explicitly capturing pairwise interactions among all image tokens, ViT effectively encodes global context and delivers competitive performance across a wide range of vision tasks~\cite{yuan2021hrformer, xie2021segformer, fang2023unleashing}. However, the quadratic cost of softmax attention incurs prohibitive computation and memory when processing high-resolution inputs, which hinders practical deployment~\cite{xie2024sana}. To mitigate this issue, many efficient attention variants have been proposed. For example, PVT~\cite{wang2021pyramid} reduces computation by spatially downsampling key–value feature maps, while Swin Transformer~\cite{liu2021swin} confines attention to local windows. While these designs improve efficiency, they also reduce the effective receptive field, weakening global context modeling and often leading to non-trivial performance drops

\begin{figure}[!t]
\centering
\includegraphics[width=0.6\columnwidth]{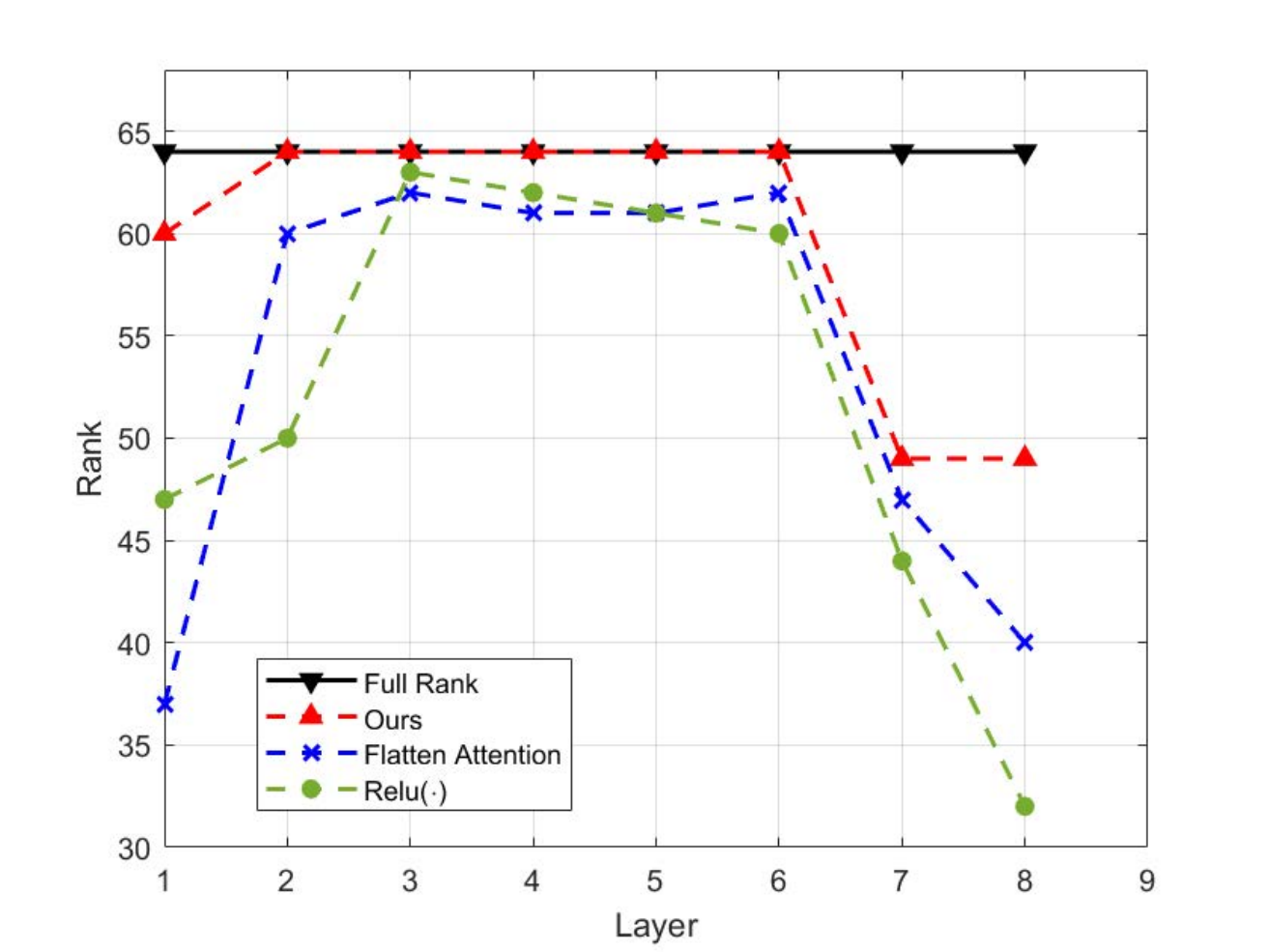}
\caption{Comparison of rank curves between $Relu\left( {{\cdot}} \right)$, Flatten Attention and our SAGA.}
\label{rank compare}
\end{figure}

To reconcile computational efficiency with a global receptive field, linear attention~\cite{katharopoulos2020transformers} has been proposed as a scalable alternative to softmax attention. By applying kernel feature mappings and reordering the computation from $(QK)V$ to $Q(KV)$, it reduces the complexity from $\mathcal{O}(N^2 d)$ to $\mathcal{O}(N d^2)$, where $N$ is the number of tokens and $d$ is the embedding dimension. This reformulation aggregates all token information into a fixed-size key–value (KV) feature map of shape $d_k \times d_v$, independent of the sequence length. The compact KV feature map can be viewed as a \textbf{global semantic repository} that encodes long-range dependencies, enabling each query to retrieve contextual information on demand~\cite{shen2021efficient}.

Despite its computational benefits, linear attention often incurs non-trivial accuracy drops relative to softmax attention. To identify the underlying cause, we systematically analyze the fixed-size $KV$ feature map—the key component that summarizes global context in linear attention. We quantify the diversity of this semantic repository via the rank profile of the $KV$ feature map. As shown in Fig.~\ref{rank compare}, across multiple linear-attention variants implemented in PVT-T, the rank trajectories remain far below the full-rank upper bound, revealing substantial redundancy in the compressed representation. This pronounced low-rank structure restricts a query’s ability to attend to diverse contextual patterns. Consequently, increasing the rank of the $KV$ feature map is critical for strengthening global context modeling and improving overall performance.

\begin{figure}[!t]
\centering
\includegraphics[width=\columnwidth]{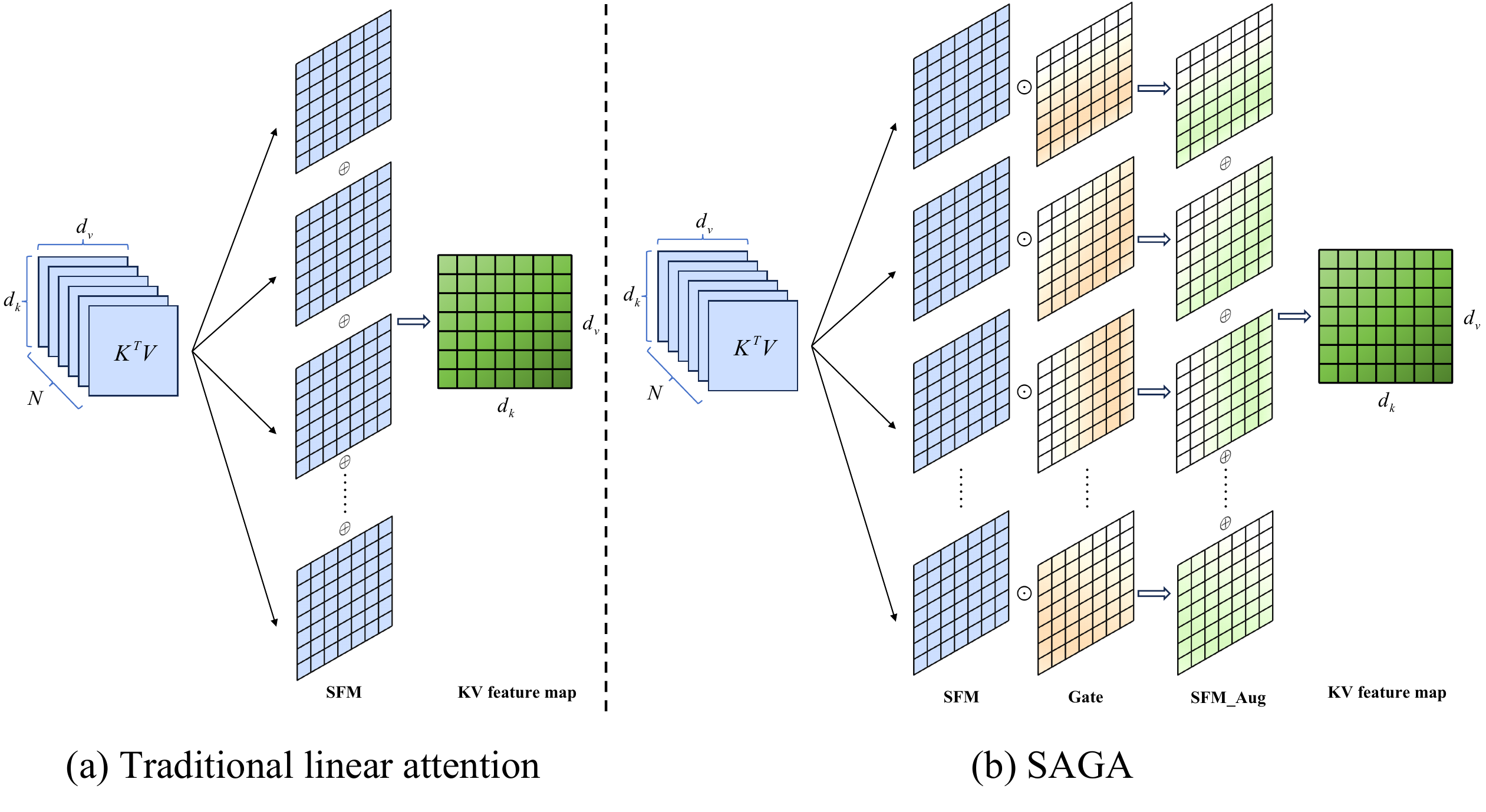}
\caption{The process of $ K^T V $ essentially involves the summation of $ N $ matrices of size $ d_k \times d_v $, which can be interpreted as the information fusion of the SFM  corresponding to $ N $ tokens. (a) Traditional linear attention aggregates all SFMs indiscriminately, which leads to substantial information redundancy. (b)We introduced a learnable gate layer for each SFM to refine the information flow entering $KV$ feature map.}
\label{GateAttention}
\end{figure}

To understand the origin of the low-rank bottleneck in the $KV$ feature map, we first examine its construction. As illustrated in Fig.~\ref{GateAttention}(a), the $KV$ feature map is formed by aggregating $N$ intermediate state feature maps (SFMs), where each SFM corresponds to a rank-1 outer product $k_i^{T} v_i$ for token $i$. This uniform, fixed-size compression merges all token-level information indiscriminately, discarding token-specific cues that are essential for accurate query discrimination. Consequently, the aggregated $KV$ feature map fails to preserve meaningful affinities between the query and the underlying key–value pairs, leading to attenuation of relevant contextual signals and amplification of irrelevant ones. We therefore attribute the low-rank structure fundamentally to this undifferentiated aggregation, which introduces substantial redundancy and constrains the representational capacity of the global semantic repository.

\begin{comment}
\textcolor{red}{As illustrated in Fig.~\ref{GateAttention} (b), RALA~\cite{fan2025breaking} assigns an input-adaptive scalar weight $\gamma$ to each SFM to modulate its importance, enabling the model to emphasize informative components while suppressing redundant information and noise. However, although scalar weighting is computationally efficient, it lacks the capacity for fine-grained information selection and filtering, and therefore provides limited control over how information flows into $S$.}
\end{comment}

To support fine-grained selection and filtering of information within SFMs—thereby reducing redundancy and increasing the expressiveness of the global semantic repository—we introduce \textbf{KVGate} module. KVGate uses learnable gates to selectively filter and aggregate token information, improving the quality of the resulting global representation. Concretely, it constructs an input-adaptive gating matrix that modulates each token-level SFM individually, as illustrated in Fig.~\ref{GateAttention}(b). This token-wise gating is particularly suitable for vision tasks without causal constraints: it controls each SFM’s contribution to the final $kv$ feature map by amplifying informative components and suppressing weak or noisy signals, enabling fine-grained token-level information flow.

However, naively incorporating gating into linear attention would require computing and storing all intermediate SFMs together with their gating matrices, incurring substantial memory overhead. To address this, we propose a \textbf{Hadamard-product decomposition} that factorizes the gates and applies them separately to the key (K) and value (V) matrices. This formulation avoids explicitly materializing intermediate SFMs and full gating matrices, reduces memory complexity, and fully exploits GPU parallelism for efficient computation.

The main contributions of this work are summarized as follows:

\begin{itemize}
\item We propose \textbf{KVGate}, a gating module for linear attention that uses input-adaptive gates to selectively modulate each intermediate SFM’s contribution to the $KV$ feature map. By amplifying informative components and suppressing uninformative ones, KVGate improves the expressiveness of the $KV$ representation.

\item We introduce a \textbf{Hadamard-product decomposition} to remove the memory bottleneck of materializing all intermediate SFMs and their gating tensors. The factorized gates are applied directly to the key (K) and value (V) matrices, eliminating explicit SFM storage, substantially reducing memory usage, and enabling efficient GPU-parallel computation.

\item Building on KVGate, we develop \textbf{SAGA} and evaluate it on image classification, semantic segmentation, object detection, and low-light image enhancement. SAGA improves ImageNet-1K Top-1 accuracy by \textbf{1.1\%} over MLLA and consistently boosts downstream tasks. On LLIE, SAGA reduces runtime and GPU memory by over \textbf{80\%} compared with LLFormer, demonstrating strong effectiveness and efficiency in practical vision settings.
\end{itemize}

\section{Related Work}
\subsection{Vision Transformer}
Following the success of Transformers and softmax-based attention in natural language processing (NLP), the vision community has shown growing interest in adapting self-attention mechanisms to visual understanding tasks. ViT~\citep{dosovitskiy2020image} pioneered this direction by dividing an image into non-overlapping patches and processing the resulting token sequence with a standard Transformer encoder. Building on ViT, DETR~\citep{carion2020end} introduced the first fully end-to-end object detector, eliminating the need for handcrafted anchor design and non-maximum suppression. SegFormer~\citep{xie2021segformer} repurposed ViT as a strong encoder for semantic segmentation, while SwinIR~\citep{liang2021swinir} employed hierarchical Transformers for image super-resolution, achieving state-of-the-art restoration quality.

Despite these advances, the quadratic complexity of self-attention with respect to the number of tokens imposes prohibitive computational and memory costs, hindering the broader deployment of ViT-based architectures in resource-constrained environments.

\subsection{Efficient Vision Transformers}
To alleviate the quadratic complexity of ViT, a variety of efficient variants have been proposed. PVT~\citep{wang2021pyramid} and DAT~\citep{xia2022vision} reduce computational costs by computing attention scores over a subset of tokens using sparse attention patterns. Swin Transformer~\citep{liu2021swin} and NAT~\citep{hassani2023neighborhood} adopt local attention mechanisms, restricting computations to confined spatial regions to mitigate overhead. Meanwhile, several hybrid architectures have been explored to combine the local inductive biases of CNNs with the global modeling capacity of Transformers~\citep{dai2021coatnet, yang2022focal, hou2024conv2former, zhou2025elaformer}, achieving competitive accuracy at reduced computational budgets.

Despite these advances, such designs do not fundamentally resolve the inherent quadratic complexity of softmax attention, which continues to limit the scalability and practicality of Transformer-based vision models.

\subsection{Linear Attention}
Linear attention fundamentally alleviates the quadratic complexity of softmax attention by replacing the softmax operator with kernel-based feature mappings and reordering the computation through the associative property of matrix multiplication to first compute $K^{T}V$~\cite{FANG2026115625, cai2026lact}. This reformulation reduces the computational complexity from $\mathcal{O}(N^2)$ to $\mathcal{O}(N)$ while maintaining global context modeling.

Existing studies on linear attention can be broadly categorized into two groups.
The first category retains a normalized attention form by redefining the similarity function. For instance, Hydra Attention~\citep{qin2022cosformer} substitutes the softmax operator with a cosine kernel, Flatten Attention~\citep{han2023flatten} applies power normalization to sharpen the attention distribution, and PolaFormer~\citep{meng2025polaformer} explicitly models query–key interactions with both identical and opposite signs to account for negative correlations. The second category abandons normalization and directly focuses on the intermediate SFMs to emphasize their relative contributions to the output. RetNet~\citep{sun2023retentive} introduces a distance-dependent decay matrix to attenuate distant SFMs, while GLA~\citep{yang2023gated} employs input-dependent gating to adaptively modulate the contribution of each SFM.

Although these approaches improve efficiency and flexibility, they still exhibit limited representational diversity due to the low-rank nature of the aggregated $KV$ feature map, leaving room for further enhancement in global context modeling.

\section{Preliminary}
\label{sec:Preliminary}
In this section, we begin by reviewing the standard softmax attention mechanism, followed by an exposition of both normalized and unnormalized variants of linear attention. For clarity of presentation, we focus on the single-head formulation.

\noindent \textbf{Softmax Attention. }Within a standard attention head, the input $x \in {{\mathbb R}^{N \times d}}$ is a sequence with token length $ N $ and dimension $ d $. Through linear mappings, $ x $ is projected into the $Q = \left\{ {{q_t}} \right\}_{t = 1}^N$, $K = \left\{ {{k_t}} \right\}_{t = 1}^N$, and $V = \left\{ {{v_t}} \right\}_{t = 1}^N$. The output $O = \left\{ {{o_t}} \right\}_{t = 1}^N$ of softmax attention can be expressed as follows:

\begin{equation}\label{eq1}
{o_t} = \frac{{\sum\nolimits_i^N {\exp \left( {{{{q_t}k_i^T} \mathord{\left/
 {\vphantom {{{q_t}k_i^T} {\sqrt d }}} \right.
 \kern-\nulldelimiterspace} {\sqrt d }}} \right)} }}{{\sum\nolimits_j^N {\exp \left( {{{{q_t}k_j^T} \mathord{\left/
 {\vphantom {{{q_t}k_j^T} {\sqrt d }}} \right.
 \kern-\nulldelimiterspace} {\sqrt d }}} \right)} }}{v_i}
\end{equation}

Through softmax attention, each token at a given position can obtain information from tokens at other positions. However, this approach incurs a computational complexity of $\mathcal{O}({N^2}d)$.

\noindent \textbf{Normalized Linear Attention. }To enhance the computational efficiency of self-attention mechanisms, linear attention replaces the softmax attention's $\exp \left( {{q_t}k_i^T} \right)$ with $\phi \left( {{q_t}} \right)\phi {\left( {{k_i}} \right)^T}$ and leverages the associative property of matrix multiplication to change the computation order from $\left( {Q{K^T}} \right)V$ to $Q\left( {{K^T}V} \right)$, thereby reducing the complexity to $\mathcal{O}(N{d^2})$. Normalized linear attention can be expressed as follows:

\begin{equation}\label{eq2}
{o_t} = \frac{{\sum\nolimits_i^N {\phi \left( {{q_t}} \right)\phi {{\left( {{k_i}} \right)}^T}{v_i}} }}{{\sum\nolimits_j^N {\phi \left( {{q_t}} \right)\phi {{\left( {{k_j}} \right)}^T}} }} = \frac{{\phi \left( {{q_t}} \right)\sum\nolimits_i^N {\phi {{\left( {{k_i}} \right)}^T}{v_i}} }}{{\phi \left( {{q_t}} \right)\sum\nolimits_j^N {\phi {{\left( {{k_j}} \right)}^T}} }}
\end{equation}

Where $\phi \left( {{\cdot}} \right)$ is a kernel function mapping that enables normalized linear attention to approximate softmax attention.

\noindent \textbf{Unnormalized Linear Attention. }In NLP field,~\cite{sun2023retentive} found that linear attention using an identity mapping kernel without normalization performs well in practice. Therefore, linear attention can be simplified as follows:

\begin{equation}\label{eq3}
{S_t} = {S_{t - 1}} + k_t^T{v_t},{\kern 1pt} {\kern 1pt} {\kern 1pt} {\kern 1pt} {\kern 1pt} {\kern 1pt} {\kern 1pt} {o_t} = {q_t}{S_t}
\end{equation}

Where ${S_t} = \sum\nolimits_{i = 1}^t {k_i^T{v_i}}  \in {{\mathbb R}^{{d_k} \times {d_v}}}$ represents the historical state, which is used to store the historical information of tokens.

\begin{figure}[!t]
\centering
\includegraphics[width=\columnwidth]{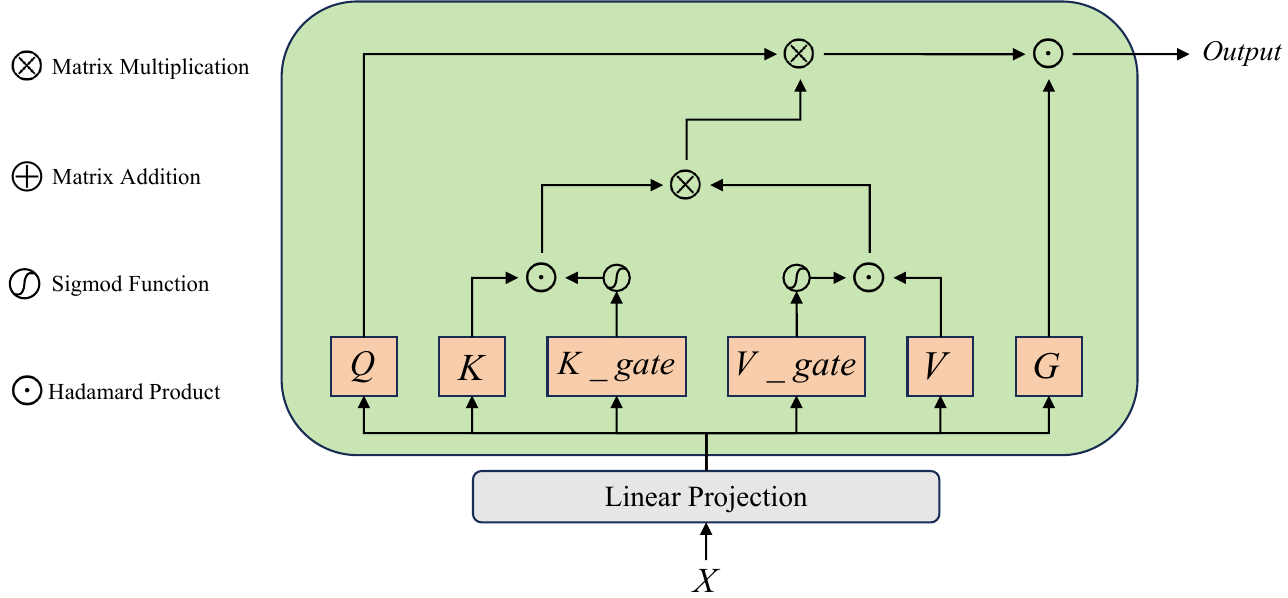}
\caption{The overall architecture of SAGA.}
\label{overall architecture}
\end{figure}

\section{Method}
\subsection{KV Gate}
Starting from Eq.~(\ref{eq3}), it is important to first distinguish the structural differences between language and vision domains. In NLP, a causal constraint is imposed, meaning that each position can only attend to tokens preceding it. As a result, the summation in Eq.~(\ref{eq3}) is computed only up to the current position $t$. In contrast, visual data does not exhibit such causality, allowing each spatial position to attend to all tokens simultaneously. Eq.~(\ref{eq3}) can thus be rewritten as follows:

\begin{equation}\label{eq6}
{o_t} = {q_t}S = {q_t}\sum\limits_{i = 1}^N {k_i^T{v_i}}
\end{equation}

Where $N$ denotes the number of tokens in an image. $k_i^T{v_i}$ represents the intermediate SFM corresponding to the $i$-th token.

When applying linear attention to visual tokens, each query $q_t$ attends to a fixed-size $KV$ feature map $S \in \mathbb{R}^{d_k \times d_v}$, from which different queries extract semantic features. Consequently, $S$ can be regarded as a \textbf{global semantic repository} comprising $d_k$ global semantic vectors, each of dimension $d_v$.

As shown in Fig.~\ref{GateAttention}(a), when the intermediate SFMs are linearly aggregated to form $S$, this undifferentiated summation fails to capture the subtle relationships among key–value pairs. As a result, $S$ is unable to emphasize relevant information or suppress irrelevant content. This limitation manifests as a low-rank structure within $S$, indicating redundancy and insufficient diversity in the global semantic repository.

To address the indiscriminate aggregation of token-wise SFMs that occurs when constructing the global semantic repository $S$ via $K^{T}V$, we introduce an input-adaptive gate $G$ that selectively modulates the contribution of each SFM during the formation of $S$. Specifically, for each token-level term $S_t=k_t^{T}v_t \in {{\mathbb R}^{{d_k} \times {d_v}}}$, we introduce a gating matrix $G_t$ matching its dimensions. $G_t$ enables fine-grained modulation of the information strength from $S_t$ flowing into the KV feature maps, thereby allowing for more precise control over the information contributed by each intermediate SFM. In visual transformer architectures, this formulation naturally supports full parallel computation across all tokens, as each gated SFM can be processed independently without temporal dependencies. Such parallelism is highly compatible with modern GPUs and significantly enhances computational efficiency while simultaneously enriching the representational capacity of $S$. Mathematically, the process corresponds to a nonlinear, weighted aggregation of all SFMs, which increases both the matrix rank and the representational diversity of the resulting $S$ feature map:

\begin{equation}\label{eq7}
{o_t} = {q_t}{S_g} = {q_t}\sum\limits_{i = 1}^N {{G_i} \odot k_i^T{v_i}}
\end{equation}

Where ${G_i} \in {\left( {0,1} \right)^{{d_k} \times {d_v}}}$ is input-adaptive gating matrix.

\subsection{Hadamard-product Decomposition}
Traditional linear attention computes $S$ directly via $K^T V$, leveraging optimized matrix multiplication implementations to minimize computational and memory costs. However, for the gating mechanism expressed in Eq.~(\ref{eq7}), the standard approach is to first compute the SFM of the N tokens together with their corresponding gating matrices, and then apply an element-wise Hadamard product for each token.

\begin{equation}
\begin{gathered}
S_i = k_i^T v_i \in \mathbb{R}^{d_k \times d_v} \\
G_i = g(k_i, v_i) \in \mathbb{R}^{d_k \times d_v} \\
\widetilde{S}_i = S_i \odot G_i
\end{gathered}
\end{equation}

This necessitates storing both the SFM and its corresponding gating tensor $G_t$ of all tokens, effectively maintaining tensors of size $2 \times N \times d_k \times d_v$.
Such storage demands lead to a substantial increase in memory consumption, which severely undermines the efficiency advantage of linear attention.

To reduce memory consumption, we derive a decomposition method of the Hadamard-product:

\paragraph{\textbf{Theorem}}
Let $u, x \in \mathbb{R}^{d \times 1}$ and $v, y \in \mathbb{R}^{1 \times d}$. Then the following identity holds:
\begin{equation}\label{Theorem}
uv \odot xy = (u \odot x)\,(v \odot y)
\end{equation}

\paragraph{\textbf{Proof}}
The matrices $uv$ and $xy$ are rank-$1$ matrices with entries
\begin{equation}
(uv)_{ij} = u_i v_j, \qquad (xy)_{ij} = x_i y_j
\end{equation}

for all $i,j \in \{1,\ldots,d\}$.

\quad \paragraph{Left-hand side}
Using the definition of the Hadamard product:
\begin{equation}
\begin{aligned}
(uv \odot xy)_{ij}
&= (uv)_{ij} (xy)_{ij} \\
&= (u_i v_j)(x_i y_j)
= (u_i x_i)(v_j y_j)
\end{aligned}
\end{equation}

\quad \paragraph{Right-hand side}
The vectors $u \odot x \in \mathbb{R}^{d \times 1}$ and $v \odot y \in \mathbb{R}^{1 \times d}$ satisfy:

\begin{equation}
(u \odot x)_i = u_i x_i, \qquad (v \odot y)_j = v_j y_j
\end{equation}

Their outer product therefore gives:

\begin{equation}
\big[(u \odot x)(v \odot y)\big]_{ij}
= (u_i x_i)(v_j y_j)
\end{equation}

\paragraph{\textbf{Conclusion}}
Since the $(i,j)$-th elements of both sides coincide, we obtain:

\begin{equation}\label{hadamad}
uv \odot xy = (u \odot x)(v \odot y)
\end{equation}

Combining Eq.(\ref{Theorem}), we decompose $G_i$ into $\alpha^T_i \beta_i$, where $\alpha_i \in {\left( {0,1} \right)^{{1} \times {d_k}}}$ and $\beta_i \in {\left( {0,1} \right)^{{1} \times {d_v}}}$:

\begin{equation}\label{eq9}
\begin{array}{l}
{o_t} = {q_t}\sum\limits_{i = 1}^N {\alpha _i^T{\beta _i} \odot k_i^T{v_i}} \\
{\kern 1pt} {\kern 1pt} {\kern 1pt} {\kern 1pt} {\kern 1pt} {\kern 1pt} {\kern 1pt} {\kern 1pt} {\kern 1pt} {\kern 1pt}  = {q_t}\sum\limits_{i = 1}^N {\left( {\alpha _i^T \odot k_i^T} \right)\left( {{\beta _i} \odot {v_i}} \right)} 
\end{array}
\end{equation}

The matrix form is as follows:

\begin{equation}\label{eq10}
O = Q\left[ {{{\left( {K \odot A} \right)}^T}\left( {V \odot B} \right)} \right]
\end{equation}

Where $A \in {\left( {0,1} \right)^{N \times {d_k}}}$ and $B \in {\left( {0,1} \right)^{N \times {d_v}}}$. 

It is necessary to additionally maintain $ A $ and $ B $, which requires storage space of $N \times \left( {{d_k} + {d_v}} \right) < 2 \times N \times {d_k} \times {d_v}$.

The generation methods for $ A $ and $ B $ are identical to those for $ Q $, $ K $ and $ V $:

\begin{equation}\label{eq11}
A = sigmoid \left( {X{W_A}} \right),B = sigmoid \left( {X{W_B}} \right)
\end{equation}

Therefore, as depicted in Fig.~\ref{overall architecture}, the final form of the linear attention proposed in this paper can be expressed as follows, where $G = X{W_g} \in {{\mathbb R}^{N \times d}}$ is introduced to further enhance nonlinearity~\cite{sun2023retentive}, $K\_gate = A$ and $V\_gate = B$:

\begin{equation}\label{eq12}
\begin{array}{*{20}{c}}
{\widetilde K = K \odot K\_gate,{\kern 1pt} {\kern 1pt} {\kern 1pt} {\kern 1pt} {\kern 1pt} {\kern 1pt} {\kern 1pt} {\kern 1pt} \widetilde V = V \odot V\_gate}\\
{O = \left[ {Q\left[ {\widetilde K^T\widetilde V} \right]} \right] \odot G}
\end{array}
\end{equation}

The introduction of this gating mechanism enables the $KV$ feature map to distinguish the importance information and remove noise from the intermediate SFM. From a mathematical perspective, this nonlinear mapping based on the Hadamard-product can significantly increase the rank of the $KV$ feature map, thereby enriching the feature diversity of the semantic repository.

\subsection{Theoretical Guarantees for Rank Enhancement}
Eq.~(\ref{hadamad}) provides theoretical guarantees for rank upper bound enhancement. 

Let $A, B \in \mathbb{R}^{m \times n}$. Each matrix can be written as a sum of rank-one terms:

\begin{equation}
\begin{array}{*{20}{c}}
A = \sum_{i=1}^r u_i v_i^T, B = \sum_{j=1}^s x_j y_j^T \\
r = \text{rank}(A), s = \text{rank}(B)\\
u_i, x_j \in \mathbb{R}^{m \times 1}, v_i, y_j \in \mathbb{R}^{n \times 1}
\end{array}
\end{equation}

Therefore: $A \odot B = \left( \sum_{i=1}^r u_i v_i^T \right) \odot \left( \sum_{j=1}^s x_j y_j^T \right) = \sum_{i=1}^r \sum_{j=1}^s \left( u_i \odot x_j \right) \left( v_i \odot y_j \right)^T$. $A \odot B$ can be viewed as a linear combination of $rs$ rank-one matrices, hence: $\text{rank}(A \odot B) \le rs = \text{rank}(A) \times \text{rank}(B)$. 

Applying this to our method, let $A, B, C, D \in \mathbb{R}^{m \times n}$. Then:

\begin{equation}
\begin{split}
\mathrm{rank}\big((A \odot C)(B \odot D)\big) &\le \min\left\{\mathrm{rank}(A \odot C), \mathrm{rank}(B \odot D)\right\}\\
&\le \min\left\{\mathrm{rank}(A) \times \mathrm{rank}(C), \mathrm{rank}(B)\times \mathrm{rank}(D)\right\}
\end{split}
\end{equation}

Consequently, our method augments the rank upper bound of KV feature maps, thereby enriching their diversity.

\subsection{Expressivity Analysis}
In this section, we provide a theoretical analysis showing that our method recovers a substantially richer order expressivity, making it strictly closer to softmax attention than the baseline linear attention. Our analysis focuses on the order structure of the output with respect to the input, which offers a clear and rigorous comparison criterion.

\noindent \textbf{Model Setup. }We analyze the limit case where the sequence length and feature dimension reduce to scalars, i.e. $n=d=1$. Let the input $X\in\mathbb{R}$. Since $Q,K,V$ are linear projections of $X$, there exist scalars
$a,b,c\in\mathbb{R}$ such that: $Q(x)=ax, K(x)=bx, V(x)=cx$. For the SAGA, the learnable gated projections reduce to scalars: $W_1=p, W_2=q$. We further define $\lambda := abc,\nu := ab$. Under this setting, the three attention operators become:

\begin{equation}
\begin{aligned}
\textbf{Baseline:}\quad
& O_0(x)=QKV
 = \lambda x^3, \\[4pt]
\textbf{SAGA:}\quad
& O_g(x)
 = Q\,(K\odot\sigma(px))\,(V\odot\sigma(qx)) \\
&\phantom{O_g(x)}
 = \lambda x^3\,\sigma(px)\sigma(qx), \\[4pt]
\textbf{Softmax:}\quad
& O_{\exp}(x)
 = e^{QK}V
 = c\,x\,e^{\nu x^2}.
\end{aligned}
\end{equation}

\noindent \textbf{Order Expressivity. }We quantify expressivity via the set of polynomial orders appearing in the local expansion.

Let $f(x)$ be analytic at $x=0$ with Taylor expansion:

\begin{equation}
f(x)=\sum_{k=0}^{\infty}\alpha_k x^k.
\end{equation}

The \emph{degree support} of $f$ is defined as:

\begin{equation}
\mathcal{S}(f):=\{\,k\in\mathbb{N}\mid \alpha_k\neq 0\,\}.
\end{equation}

For a model family $\mathcal{F}$, its expressible degree support is:

\begin{equation}
\mathcal{S}(\mathcal{F}) := \bigcup_{f\in\mathcal{F}} \mathcal{S}(f).
\end{equation}

\textbf{Baseline. }For the baseline linear attention family $\mathcal{F}_0=\{\,\lambda x^3 \mid \lambda\in\mathbb{R}\,\}$, the expressible degree support satisfies

\begin{equation}
\mathcal{S}(\mathcal{F}_0)=\{3\}.
\end{equation}

This indicates that the baseline linear attention mechanism is restricted to a single cubic polynomial form and is incapable of modeling higher-order interactions.

\textbf{Softmax. }For the softmax family $\mathcal{F}_{\exp}
=\left\{\, c\,x\,e^{\nu x^2} \;\middle|\; c,\nu\in\mathbb{R} \,\right\}$, when $\nu\neq 0$, the exponential term admits the Taylor expansion:

\begin{equation}
e^{\nu x^2}
=\sum_{m=0}^{\infty}\frac{\nu^m}{m!}x^{2m}.
\end{equation}

Multiplying by $cx$ yields:
\begin{equation}
cx\,e^{\nu x^2}
=\sum_{m=0}^{\infty} c\,\frac{\nu^m}{m!}x^{2m+1}.
\end{equation}

Thus:

\begin{equation}
\mathcal{S}(\mathcal{F}_{\exp}) \supseteq \{1,3,5,7,\dots\}.
\end{equation}

This indicates that softmax inherently possesses an infinite hierarchy of expressive structures (spanning all odd orders), enabling it to generate a rich family of complex functions and thus endowing it with exceptionally strong expressive power.

\textbf{SAGA. }For the SAGA family $\mathcal{F}_g=\left\{\, \lambda x^3 \sigma(px)\sigma(qx)|\lambda,p,q \in \mathbb{R} \,\right\}$, it admits a function family with the same order richness as softmax attention. Consequently, SAGA achieves significantly enhanced expressive power while maintaining linear computational complexity.

Choose symmetric gates $q=-p$. Using $\sigma(-z)=1-\sigma(z)$, we obtain:

\begin{equation}
\sigma(px)\sigma(-px)
=\sigma(px)\bigl(1-\sigma(px)\bigr)
=\sigma'(px)
\end{equation}

Where $\sigma'(z)$ is an even, analytic, non-polynomial function. Hence:

\begin{equation}
\sigma'(px)
= \sum_{m=0}^{\infty} \beta_m x^{2m},
\qquad \beta_0 \neq 0.
\end{equation}

Multiplying by the outer factor $x^3$ yields

\begin{equation}
O_g(x)
= \lambda x^3 \sum_{m=0}^{\infty} \beta_m x^{2m}
= \sum_{m=0}^{\infty} (\lambda \beta_m) x^{2m+3}.
\end{equation}

So:

\begin{equation}
\mathcal{S}(\mathcal{F}_{\exp}) \supseteq \{3,5,7,9,\dots\}.
\end{equation}

\textbf{Main Result. }In summary, the supports of the expressible degrees satisfy:

\begin{equation}
\mathcal{S}(\mathcal{F}_0)
\;\subset\;
\mathcal{S}(\mathcal{F}_g)
\;\subset\;
\mathcal{S}(\mathcal{F}_{\exp})
\end{equation}

Consequently, SAGA admits an infinite odd-degree expansion that aligns with the order structure induced by softmax attention, whereas the baseline linear attention is restricted to a single cubic term. Therefore, in terms of order expressivity, the SAGA is strictly closer to softmax attention than the baseline linear attention.

\subsection{Complexity Analysis}
Let $d$ denote the number of channels, and $k$ denote the size of the convolution kernel. The computational cost of calculating $ Q $, $ K $, $ V $, $ K\_gate $, $ V\_gate $, $ G $ and outputs projections is $ 7Nd^2 $. The Hadamard-product of $(K, K\_gate)$ and $(V, V\_gate)$ has a computational cost of $2Nd$. The attention computation for $(Q, K, V)$ requires a computational cost of $2Nd^2$. The convolution operation requires $k^2Nd$. The computational cost required for output augment $G$ is $Nd$. In summary, the total complexity of SAGA is given by Eq.(\ref{supp_eq13}), and it exhibits linear complexity with respect to $N$.

\begin{equation}\label{supp_eq13}
\Theta  = 9N{d^2} + {k^2}Nd + 3Nd
\end{equation}

\begin{table}[h]
\centering
\scalebox{0.9}{\begin{tabular}{c|c|c|c}
\toprule
\textbf{Model} & Blocks & Channels & Heads \\
\midrule
SAGA-T & {[}2, 2, 6, 2{]} & {[}64, 128, 256, 512{]} & {[}1, 2, 4, 8{]} \\
SAGA-S & {[}3, 5, 9, 3{]} & {[}64, 128, 320, 512{]} & {[}1, 2, 5, 8{]} \\
SAGA-B & {[}4, 6, 12, 6{]} & {[}96, 192, 384, 512{]} & {[}1, 2, 6, 8{]} \\
SAGA-L & {[}4, 7, 19, 8{]} & {[}96, 192, 448, 640{]} & {[}1, 2, 7, 10{]} \\
\bottomrule
\end{tabular}}
\caption{Architecture details of SAGA.}
\label{Architecture details}
\end{table}

\section{Experimental setup}
We validated the effectiveness of the proposed model on the tasks of image classification, semantic segmentation and object detection using the ImageNet-1K~\cite{deng2009imagenet}, ADE20K~\cite{zhou2019semantic} and COCO datasets~\cite{lin2014microsoft}. First of all, We train the models from scratch on ImageNet-1K. Subsequently, using the pre-trained models as backbones, we fine-tuned image segmentation models on ADE20K and object detection models on COCO. To validate the performance and efficiency of our method in long-context tasks, we conducted experiments on low-light image enhancement using the LOL~\cite{wei2018deep} and the MIT-Adobe FiveK~\cite{bychkovsky2011learning} as the dataset. All models are trained and fine-tuned on 8 NVIDIA RTX 3090 and 4090 GPUs. Additionally, comprehensive ablation studies are conducted to analyze the effectiveness of the components.

\begin{table}[!t]
  \centering
  % 主表标题（统一编号）

  % 左侧子表
  \begin{minipage}[t]{0.46\columnwidth}
    \raggedleft
    \scalebox{0.75}{\begin{tabular}{l|ccc}
    \toprule
    MODEL & Para & FLOPs & Acc \\
    \midrule
    PVTv2-b1~\cite{wang2022pvt}   & 13M & 2.1G & 78.7 \\
    VAN-b1~\cite{guo2023visual}       & 14M & 2.5G & 81.1 \\
    Conv2Former-N~\cite{hou2024conv2former} & 15M & 2.2G & 81.5 \\
    SBCFormer-L~\cite{lu2024sbcformer}   & 19M & 2.7G & 81.1 \\
    FLattn-PVT-T~\cite{han2023flatten}  & 11M & 1.9G & 77.8 \\
    Inline-PVT-T~\cite{han2024bridging} & 12M & 2.0G & 78.2 \\
    Agent-PVT-T~\cite{han2024agent}  & 12M & 2.0G & 78.4 \\
    Pola-PVT-T~\cite{meng2025polaformer}   & 12M & 2.0G & 78.8 \\
    RAVLT-T~\cite{fan2025breaking}      & 15M & 2.4G & $82.3^*$ \\
    MAViT-T~\cite{fan2025rectifying}         & 16M & 2.5G & $82.4^*$ \\
    \cellcolor{gray!30}SAGA-T & \cellcolor{gray!30}15M & \cellcolor{gray!30}2.7G & \cellcolor{gray!30}\textbf{82.8} \\
    \midrule
    Conv2Former-T~\cite{hou2024conv2former} & 27M & 4.4G & 83.2 \\
    MambaOut-T~\cite{yu2025mambaout}    & 27M & 4.5G & 82.7 \\
    InternImage-T~\cite{wang2023internimage}& 30M & 5.0G & 83.5 \\
    Vim-S~\cite{zhu2024vision}         & 26M & 3.7G & 80.6 \\
    VMamba-T~\cite{liu2024vmamba}      & 30M & 4.9G & 82.6 \\
    LocalVMamba-T~\cite{huang2024localmamba}& 26M & 5.7G & 82.7 \\
    VRWKV-S~\cite{duan2024vision}      & 24M & 4.6G & 80.1 \\
    ViG-H-T~\cite{liao2025vig}      & 29M & 4.5G & 82.8 \\
    FLatten-Swin-T~\cite{han2023flatten}& 29M & 4.5G & $82.1^*$ \\
    Inline-Swin-T~\cite{han2024bridging} & 30M & 4.5G & $82.4^*$ \\
    Agent-Swin-T~\cite{han2024agent} & 29M & 4.5G & 82.6 \\
    Pola-Swin-T~\cite{meng2025polaformer}  & 29M & 4.5G & 82.6 \\
    MLLA-T~\cite{han2024demystify}       & 25M & 4.2G & 83.5 \\
    \cellcolor{gray!30}SAGA-S & \cellcolor{gray!30}26M & \cellcolor{gray!30}5.1G & \cellcolor{gray!30}\textbf{84.4} \\
    \bottomrule
  \end{tabular}}
  \end{minipage}
  \hspace{1pt}
  % 右侧子表
  \begin{minipage}[t]{0.46\columnwidth}
    \raggedright
    \scalebox{0.75}{\begin{tabular}{l|ccc}
    \toprule
    MODEL & Para & FLOPs & Acc \\
    \midrule
    MambaOut-S~\cite{yu2025mambaout}    & 49M & 9.0G & 84.1 \\
    MogaNet-B~\cite{li2022moganet}      & 44M & 9.9G & 84.3 \\
    ViG-H-S~\cite{liao2025vig}      & 50M & 8.8G & 83.8 \\
    VMamba-S~\cite{liu2024vmamba}      & 50M & 8.7G & 83.6 \\
    FLattn-Swin-S~\cite{han2023flatten} & 51M & 8.7G & 83.5 \\
    Inline-Swin-S~\cite{han2024bridging} & 50M & 8.7G & 83.6 \\
    Agent-Swin-S~\cite{han2024agent} & 50M & 8.7G & 83.7 \\
    Pola-Swin-S~\cite{meng2025polaformer}  & 50M & 8.7G & 83.6 \\
    MLLA-S~\cite{han2024demystify}       & 43M & 7.3G & 84.4 \\
    SOFT-L~\cite{lu2021soft}        & 64M & 11G  & 83.1 \\
    \cellcolor{gray!30}SAGA-B & \cellcolor{gray!30}49M & \cellcolor{gray!30}11G & \cellcolor{gray!30}\textbf{85.3} \\
    \midrule
    Swin-B~\cite{liu2021swin}         & 88M & 15G  & 83.3 \\
    InterImage-B~\cite{wang2023internimage} & 97M & 16G  & 84.9 \\
    MambaOut-S~\cite{yu2025mambaout}    & 85M & 16G  & 84.2 \\
    VMamba-B~\cite{liu2024vmamba}      & 89M & 15G  & 83.9 \\
    ViG-H-B~\cite{liao2025vig}      & 89M & 16G  & 84.2 \\
    SMT-L~\cite{lin2023scale}         & 81M & 18G  & 84.6 \\
    SG-Former-B~\cite{ren2023sg}      & 78M & 16G  & 84.7 \\
    FLatten-Swin-B~\cite{han2023flatten}& 89M & 15G  & 83.8 \\
    Inline-Swin-B~\cite{han2024bridging}& 88M & 15G  & 84.1 \\
    Agent-Swin-B~\cite{han2024agent} & 88M & 15G  & 84.0 \\
    Pola-Swin-B~\cite{meng2025polaformer}  & 88M & 15G  & 83.8 \\
    MLLA-B~\cite{han2024demystify}       & 96M & 16G  & 85.3 \\
    VRWKV-B~\cite{duan2024vision}      & 94M & 18G  & 82.0 \\
    \cellcolor{gray!30}SAGA-L & \cellcolor{gray!30}96M & \cellcolor{gray!30}18G & \cellcolor{gray!30}\textbf{85.5} \\
    \bottomrule
  \end{tabular}}
  \end{minipage}
  \caption{Comparison of image classification results on the ImageNet-1K dataset. Results marked with an * are reproduced under the same training setup as the original paper.}
  \label{classification result} % 主表标签，用于引用
\end{table}

\subsection{Image Classification}

\noindent \textbf{Model Architecture. }
Based on~\cite{fan2025breaking}, we constructed a hierarchical vision backbone and categorized the models into four groups according to parameter and FLOPs, with the specific configurations of block numbers, channel dimensions and attention heads detailed in Tab. \ref{Architecture details}.

\noindent \textbf{Settings. }We trained the model on ImageNet-1K for the classification task, which comprises 1.28M training images and 50K validation images. We employed Top-1 accuracy as the evaluation metric and compared our model with other SOTA models. We trained all models using the Adam~\cite{loshchilov2017decoupled} optimizer with a cosine learning rate decay over 300 epochs and a linear warm-up for 20 epochs. When the batch size is 1024, the base learning rate is set to \(2 \times 10^{-3}\) and then scaled linearly with the batch size. The weight decay is set to 0.05 and data augmentation is employed to prevent overfitting.

\noindent \textbf{Results. }As shown in Tab.~\ref{classification result}, the proposed model demonstrates consistent and substantial improvements over baseline approaches. These results indicate that SAGA effectively captures informative token interactions, thereby enhancing the representational capacity of visual feature maps.

\begin{figure}[t]
    \centering
    \scalebox{0.8}{\begin{subfigure}[t]{0.32\linewidth}
        \centering
        \includegraphics[width=\linewidth]{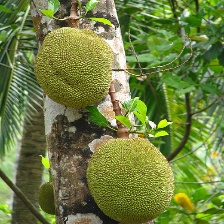}
        \caption{Original}
    \end{subfigure}
    \hfill
    \begin{subfigure}[t]{0.32\linewidth}
        \centering
        \includegraphics[width=\linewidth]{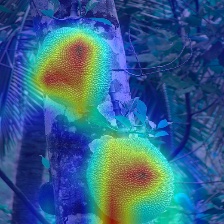}
        \caption{SAGA}
    \end{subfigure}
    \hfill
    \begin{subfigure}[t]{0.32\linewidth}
        \centering
        \includegraphics[width=\linewidth]{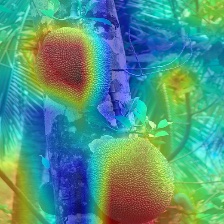}
        \caption{Linear Attention}
    \end{subfigure}}
    \caption{AblationCAM of SAGA and Linear Attention.}
    \label{AblationCAM}
\end{figure}

\noindent \textbf{Visualization. }As illustrated in Fig.~\ref{AblationCAM}, we visualize the proposed SAGA using Ablation-CAM~\cite{ramaswamy2020ablation} and compare it with standard linear attention. The results show that our method more consistently attends to semantically relevant regions while effectively suppressing irrelevant responses, demonstrating enhanced contextual modeling capability.

\begin{table}[!t]
\centering
\resizebox{\linewidth}{!}{% 自适应表格宽度
\begin{tabular}{l|cc|cccccc|cccccc}
\toprule
\multirow{2}{*}{Method} & \multirow{2}{*}{\makecell{Para\\(M)}} & \multirow{2}{*}{\makecell{FLOPs\\(G)}} & \multicolumn{6}{c|}{Mask R-CNN 1$\times$} & \multicolumn{6}{c}{Mask R-CNN 3$\times$} \\
& & & AP$^b$ & AP$^b_{50}$ & AP$^b_{75}$ & AP$^m$ & AP$^m_{50}$ & AP$^m_{75}$ & AP$^b$ & AP$^b_{50}$ & AP$^b_{75}$ & AP$^m$ & AP$^m_{50}$ & AP$^m_{75}$ \\
\midrule
PVT-T~\cite{wang2021pyramid} & 33 & 240 & 36.7 & 59.2 & 39.3 & 35.1 & 56.7 & 37.3 & 39.8 & 62.2 & 43.0 & 37.4 & 59.3 & 39.9 \\
SOFT++-Tiny~\cite{lu2021soft} & 32 & 212 & 41.2 & 63.7 & 44.7 & 38.2 & 61.0 & 41.0 & -- & -- & -- & -- & -- & -- \\
MPVIT-T~\cite{lee2022mpvit} & 28 & 216 & 42.2 & 64.2 & 45.8 & 39.0 & 61.4 & 41.8 & 44.8 & 66.9 & 49.2 & 41.0 & 64.2 & 44.1 \\
RAVLT-T~\cite{fan2025breaking} & 33 & 219 & 47.2 & 69.1 & 51.7 & 42.5 & 66.0 & 46.0 & -- & -- & -- & -- & -- & -- \\
\rowcolor{gray!30} SAGA-T & 34 & 221 & \textbf{47.2} & \textbf{69.1} & \textbf{52.0} & \textbf{42.7} & \textbf{66.3} & \textbf{46.0} & \textbf{48.7} & \textbf{70.3} & \textbf{53.5} & \textbf{43.4} & \textbf{67.1} & \textbf{46.6} \\
\midrule
MPVIT-S~\cite{lee2022mpvit} & 43 & 268 & 46.4 & 68.6 & 51.2 & 42.4 & 65.6 & 45.7 & 48.4 & 70.5 & 52.6 & 43.9 & 67.6 & 47.5 \\
FL-Swin-T~\cite{han2023flatten} & 49 & 268 & 46.5 & 66.1 & 47.9 & 40.2 & 63.1 & 43.0 & 46.5 & 68.5 & 50.8 & 42.1 & 65.4 & 45.1 \\
CSWin-T~\cite{dong2022cswin} & 42 & 279 & 46.7 & 68.6 & 51.3 & 42.2 & 65.6 & 45.4 & 49.0 & 70.7 & 53.7 & 43.6 & 67.9 & 46.6 \\
VMamba-T~\cite{liu2024vmamba} & 50 & 271 & 47.3 & 69.3 & 52.0 & 42.7 & 66.4 & 45.9 & 48.8 & - & - & 43.7 & - & - \\
MLLA-T~\cite{han2024demystify} & 44 & 255 & 46.8 & 69.5 & 51.5 & 42.1 & 66.4 & 45.0 & 48.8 & 71.0 & 53.6 & 43.8 & 68.0 & 46.8 \\
InternImage-T~\cite{wang2023internimage} & 49 & 270 & 47.2 & 69.0 & 52.1 & 42.5 & 66.1 & 45.8 & 49.1 & 70.4 & 54.1 & 43.7 & 67.3& 47.3 \\
NaLaFormer-S~\cite{meng2025nalaformer} & 44 & 272 & 49.5 & 71.2 & 54.3 & 44.2 & 68.1 & 47.8 & 49.7 & 70.5 & 54.7 & 44.3 & 68.0 & 48.0 \\
\rowcolor{gray!30} SAGA-S & 44 & 266 & \textbf{50.1} & \textbf{71.4} & \textbf{55.3} & \textbf{44.7} & \textbf{68.3} & \textbf{48.4} & \textbf{51.0} & \textbf{71.9} & \textbf{55.9} & \textbf{45.1} & \textbf{68.5} & \textbf{48.9} \\
\bottomrule
\end{tabular}
}
\caption{Object detection and instance segmentation results on the COCO dataset using Mask R-CNN with $1\times$ and $3\times$ schedule.}
\label{mask_rcnn_coco}
\end{table}

\begin{table}[!t]
\centering
\setlength{\tabcolsep}{2pt}
\renewcommand{\arraystretch}{1.0}
\scalebox{0.75}{\begin{tabular}{c|cc|cccccc}
\toprule
\multirow{2}{*} & \multirow{2}{*}{\makecell{Para\\(M)}} & \multirow{2}{*}{\makecell{FLOPs\\(G)}} & \multicolumn{6}{c}{RetinaNet 1$\times$} \\
& & & $AP^b$ & $AP^b_{50}$ & $AP^b_{75}$ & $AP^b_{S}$ & $AP^b_{M}$ & $AP^b_{L}$ \\
\toprule
PVTv2-B1~\cite{wang2022pvt} & 23 & 225 & 41.2 & 61.9 & 43.9 & 25.4 & 44.5 & 54.3 \\
MPViT-XS~\cite{lee2022mpvit} & 20 & 211 & 43.8 & 65.0 & 47.1 & 28.1 & 47.6 & 56.5 \\
SOFT++-Tiny~\cite{lu2021soft} & 23 & 200 & 41.9 & 62.7 & 44.7 & 27.8 & 45.4 & 55.6 \\
Pola-PVT-T~\cite{meng2025polaformer} & - & - & 40.0 & 60.7 & 42.7 & 25.0 & 43.6 & 52.9 \\
Agent-PVT-T~\cite{han2024agent} & - & 211 & 40.3 & 61.2 & 42.9 & 25.5 & 43.4 & 54.3 \\
RMT-T~\cite{fan2024rmt} & 23 & 199 & 45.1 & 66.2 & 48.1 & 28.8 & 48.9 & 61.1 \\
\rowcolor{gray!30} SAGA-T & 24 & 203 & \textbf{46.2} & \textbf{67.8} & \textbf{49.5} & \textbf{29.6} & \textbf{50.0} & \textbf{61.6} \\
\midrule
Swin-T~\cite{liu2021swin} & 38 & 248 & 41.7 & 63.1 & 44.3 & 27.0 & 45.3 & 54.7 \\
CMT-S~\cite{guo2022cmt} & 44 & 231 & 44.3 & 65.5 & 47.5 & 27.1 & 48.3 & 59.1 \\
CrossFormer-S~\cite{wang2023crossformer++} & 41 & 272 & 44.4 & 65.8 & 47.4 & 28.2 & 48.4 & 59.4 \\
MPViT-XS~\cite{lee2022mpvit} & 32 & 248 & 45.7 & 57.3 & 48.8 & 28.7 & 49.7 & 59.2 \\
Pola-PVT-T~\cite{meng2025polaformer} & - & - & 43.2 & 64.1 & 46.4 & 28.0 & 46.4 & 57.9 \\
Agent-PVT-S~\cite{han2024agent} & - & 274 & 44.1 & 65.3 & 47.3 & 29.2 & 47.5 & 59.8 \\
\rowcolor{gray!30} SAGA-S & 34 & 248 & \textbf{48.3} & \textbf{69.6} & \textbf{52.2} & \textbf{30.9} & \textbf{52.6} & \textbf{63.8} \\
\bottomrule
\end{tabular}}
\caption{Comparison of object detection and instance segmentation results on the COCO datasets using RetinaNet with 1 $\times$ schedule.}
\label{RetinaNet 1x}
\end{table}

\subsection{Object Detection}
\noindent \textbf{Settings. }We employed SAGA-T and SAGA-S, both pre-trained with ImageNet-1K weights, as backbones to train the Mask-RCNN~\cite{he2017mask} and RetinaNet~\cite{lin2017focal} frameworks on the COCO dataset. Each schedule comprises 12 training epochs and the AdamW optimizer is applied with a learning rate of $1e-4$ and a weight decay of $1e-4$. SAGA-RetinaNet is trained under the 1x schedule setting, SAGA-MRCNN is trained under both the 1x and 3x schedule settings.

\noindent \textbf{Results. }As shown in Tab.~\ref{mask_rcnn_coco} and~\ref{RetinaNet 1x}, the proposed model consistently achieves superior performance across all detection frameworks. When trained within the RetinaNet framework, SAGA-T and SAGA-S reach 46.2\% and 48.3\% AP$^b$, respectively. Under the Mask R-CNN framework with a 3× training schedule, the same backbones obtain 48.7\% and 51.0\% AP$^b$, surpassing all baseline models. These results demonstrate that SAGA effectively extracts task-relevant information from intermediate SFMs while suppressing redundant content, thereby enhancing the expressive capacity of the $KV$ feature maps and improving overall detection performance.

\begin{table}[!t]
\centering
\scalebox{0.75}{\begin{tabular}{c|ccc|ccc}
\hline
\multirow{2}{*}{Model} & \multicolumn{3}{c|}{Semantic FPN 80K} & \multicolumn{3}{c}{Upernet 160K} \\
 & Para & FLOPs & mIoU & Para & FLOPs & mIoU \\
\hline
VAN-B1~\cite{guo2023visual} & 18M & 140G & 42.9 & -- & -- & -- \\
PVTv2-B1~\cite{wang2022pvt} & 18M & 136G & 42.5 & -- & -- & -- \\
RMT-T~\cite{fan2024rmt} & 17M & 136G & 46.4 & -- & -- & -- \\
NaLaFormer-T~\cite{meng2025nalaformer} & -- & -- & 45.4 & -- & -- & 46.9 \\
\rowcolor{gray!20} SAGA-T & 18M & 138G & \textbf{47.7} & 44M & 895G & \textbf{48.1} \\
\hline
MogaNet-S~\cite{li2022moganet} & 29M & 189G & 47.7 & 55M & 946G & 49.2 \\
StructViT-S~\cite{kim2024learning} & 26M & 271G & 46.9 & -- & -- & -- \\
SMT-S~\cite{lin2023scale} & -- & -- & -- & 50M & 935G & 49.2 \\
DAT-T~\cite{xia2022vision} & 32M & 198G & 42.6 & 60M & 957G & 45.5 \\
RMT-S~\cite{fan2024rmt} & 30M & 180G & 49.4 & 56M & 937G & 49.8 \\
NaLaFormer-S~\cite{meng2025nalaformer} & -- & -- & 48.0 & -- & -- & 48.5 \\
RAVLT-S~\cite{fan2025breaking} & 28M & 180G & 49.5 & 55M & 937G & 50.7 \\
\rowcolor{gray!20} SAGA-S & 28M & 184G & \textbf{50.8} & 55M & 941G & \textbf{51.3} \\
\hline
\end{tabular}}
\caption{Comparison of semantic segmentation results on the ADE20K datasets using Semantic FPN and Upernet.}
\label{Semantic FPN and Upernet}
\end{table}

\subsection{Semantic Segmentation}
\noindent \textbf{Settings. }We employed SAGA-T and SAGA-S, both pre-trained with ImageNet-1K weights, as backbones to train the SemanticFPN~\cite{kirillov2019panoptic} and UperNet~\cite{xiao2018unified} frameworks on the ADE20K dataset. The SemanticFPN models are configured with 80K training iterations, while the UperNet models use 160K iterations, employing the AdamW optimizer with a learning rate of $6e-5$. The weight decay for SemanticFPN is set to $1e-3$, whereas that for UperNet is configured as $1e-2$.

\noindent \textbf{Results. }As shown in Tab.~\ref{Semantic FPN and Upernet}, SAGA demonstrates superior segmentation performance under comparable parameter and FLOPs. Specifically, SAGA-T and SAGA-S achieve mIoU scores of 47.7\% and 50.8\% respectively under the Semantic FPN framework, and 48.1\% and 51.3\% under the UperNet framework, surpassing other baseline models.

\begin{table}[!t]
    \centering
    \setlength{\tabcolsep}{1pt}
    \scalebox{0.75}{
    \begin{tabular}{c|ccc|ccc|ccc}
    \toprule
    \multirow{2}{*}{Methods} & \multirow{2}{*}{\makecell{Para\\(M)}} & \multirow{2}{*}{\makecell{RT\\(ms)}} & \multirow{2}{*}{\makecell{Mem\\(M)}} & \multicolumn{3}{c|}{\textbf{LOL}} & \multicolumn{3}{c}{\textbf{MIT-Adobe FiveK}} \\
     & & & & PSNR $\uparrow$ & SSIM $\uparrow$ & LPIPS $\downarrow$ & PSNR $\uparrow$ & SSIM $\uparrow$ & LPIPS $\downarrow$ \\
    \midrule
    DSLR~\cite{lim2020dslr}            & -- & -- & -- & 14.98 & 0.596 & 0.376 & 20.24 & 0.829 & 0.153 \\
    Z\_DCE~\cite{guo2020zero}            & -- & -- & -- & 14.86 & 0.562 & 0.335 & 15.93 & 0.767 & 0.165 \\
    Z\_DCE++~\cite{li2021learning}    & -- & -- & -- & 14.75 & 0.518 & 0.328 & 14.61 & 0.405 & 0.231 \\
    RUAS~\cite{liu2021retinex}             & -- & -- & -- & 16.40 & 0.503 & 0.270 & 15.99 & 0.786 & 0.139 \\
    ELGAN~\cite{jiang2021enlightengan}           & -- & -- & -- & 17.48 & 0.652 & 0.322 & 17.91 & 0.836 & 0.143 \\
    Restormer~\cite{zamir2022restormer}       & -- & -- & -- & 22.37 & 0.816 & \textbf{0.141} & 24.92 & 0.911 & 0.058 \\
    LLFormer~\cite{wang2023ultra}                   & 24.6 & 4631.71 & 33106.9 & 23.65 & \textbf{0.816} & 0.169 & \textbf{25.75} & \textbf{0.923} & \textbf{0.045} \\
    SAGA                   & 22.4 & 884.64 & 6237.7 & \textbf{23.99} & 0.809 & 0.171 & 25.60 & 0.922 & 0.047 \\
    \bottomrule
    \end{tabular}
    }
    \caption{Comparison results on LOL and MIT-Adobe FiveK datasets in terms of PSNR, SSIM, LPIPS between our method and other models on low light image enhancement. The ``RT'' denotes the inference time and ``Mem'' denotes peak GPU memory. All reported data are measured at a resolution of 1568×1568 using a single RTX 4090 GPU. }
    \label{low light image enhancement}
\end{table}

\begin{figure}[!t]
    \centering
    \includegraphics[width=0.9\textwidth]{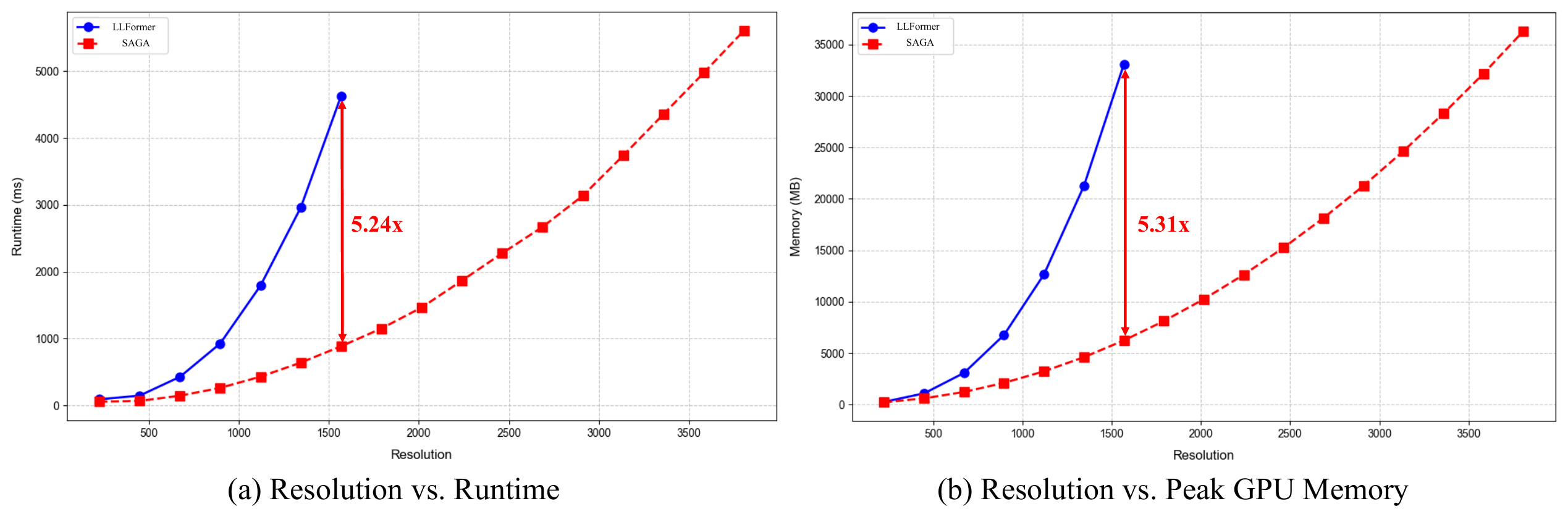}
    \caption{Efficiency Comparison between LLFormer and SAGA. All reported data are measured using a single RTX 4090 GPU.}
    \label{LLFormer runtime and peak gpu memory}
\end{figure}

\begin{figure}[!t]
\centering
\includegraphics[width=\textwidth]{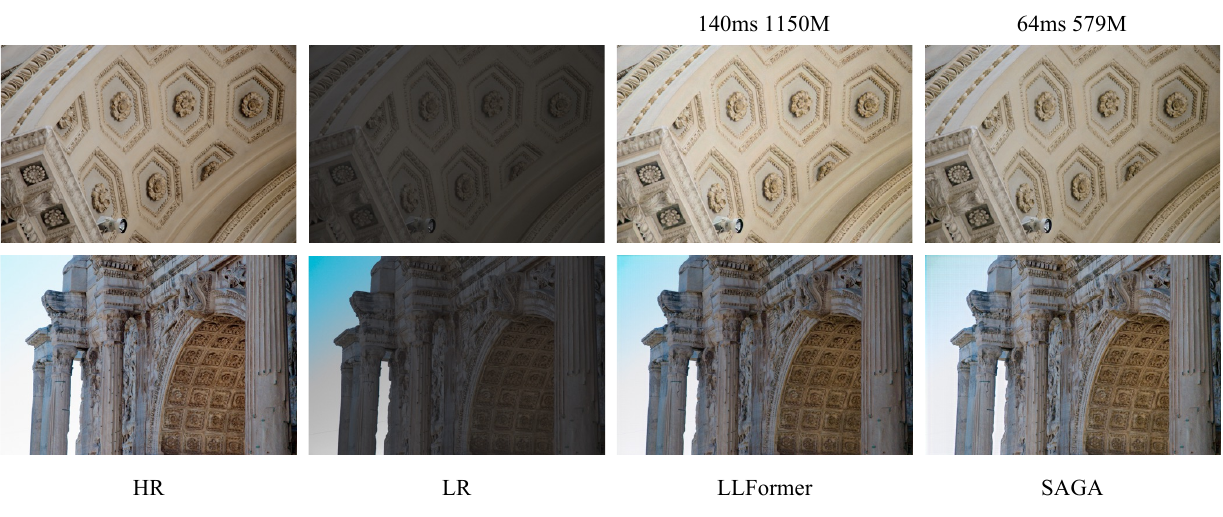}
\caption{Visualizations illustrating SAGA’s low-light image enhancement results on the MIT-Adobe FiveK dataset. Runtime and Peak GPU Memory measurements are obtained on a single NVIDIA RTX 4090 GPU. }
\label{LLIECompare}
\end{figure}

\subsection{Low Light Image Enhancemen}
\noindent \textbf{Settings. }To validate the effectiveness of the proposed method in long-context tasks, we conducted low-light image enhancement experiments building upon LLFormer. The models are trained using the LOL (240k tokens) and MIT-Adobe FiveK (262k tokens) datasets, during which the Axis-based Multi-head Self-Attention in LLFormer is replaced. Performance is evaluated using PSNR, SSIM and LPIPS metrics, while inference runtime and peak GPU memory usage are recorded. The detailed configurations can be found in the supplementary material. We only replace the a-module in the LLFormer architecture. The model is trained on 128×128 image patches. For data augmentation, we apply both horizontal and vertical flipping. We adopt the Adam optimizer with an initial learning rate of $1e-4$, which is gradually decayed to $1e-6$ using cosine annealing.

\noindent \textbf{Results. }As shown in Table~\ref{low light image enhancement}, SAGA achieves competitive performance across PSNR, SSIM, and LPIPS metrics compared to all baseline methods. Although its accuracy is marginally lower than that of LLFormer, it demonstrates a significant improvement in efficiency. At a resolution of 1568×1568, it reduces runtime by 80.9\% and peak GPU memory usage by 81.2\%, while also maintaining a smaller parameter count. Furthermore, as illustrated in Fig.~\ref{LLFormer runtime and peak gpu memory} and Fig.~\ref{LLIECompare}, SAGA scales linearly in both runtime and memory consumption as image resolution increases, enabling the processing of high-resolution inputs with stable efficiency.

\subsection{Ablation Study}
% Table generated by Excel2LaTeX from sheet 'Sheet7'
\begin{table}[!t]
  \centering
    \begin{tabular}{c|c|c|c}
    \toprule
    Gate  & Para(M) & FLOPs(G) & Acc(\%) \\
    \midrule
    Self sigmoid & 26    & 4.2   & 82.7 \\
    Low Rank & 27    & 4.7     & 82.8 \\
    \midrule
    Polaformer & 29    & 4.5   & 82.6 \\
    SAGA  & 28    & 4.6   & \textbf{83.0} \\
    \bottomrule
    \end{tabular}%
    \caption{More Analysis About Gate}
  \label{AnalysGate}%
\end{table}%

\noindent \textbf{Analysis About Gate. }Since the method introduces additional parameters in the generation of the gate, to further verify the effectiveness of the proposed gating mechanism, we conducted comparative experiments using two alternative methods for generating the gate.

\textbf{Self sigmoid. } This method no longer generates a separate gating matrix but instead performs self-enhancement by applying the sigmoid operation to $K$ and $V$. The formula is shown as follows:

\begin{equation}\label{eq14}
\begin{array}{*{20}{c}}
{\widetilde K = K \odot sigmoid \left( K \right) ,{\kern 1pt} {\kern 1pt} {\kern 1pt} {\kern 1pt} {\kern 1pt} {\kern 1pt} {\kern 1pt} {\kern 1pt} \widetilde V = V \odot sigmoid \left( V \right) }\\
{O = \left[ {Q\left[ {\widetilde K^T\widetilde V} \right]} \right] \odot G}
\end{array}
\end{equation}

\textbf{Low Rank. } This method obtains $K\_gate$ and $V\_gate$ through low-rank matrices to reduce the number of introduced parameters. The formulas are shown as follows:

\begin{equation}\label{eq15}
\begin{array}{*{20}{c}}
{K \_gate = sigmoid \left( {X{W_k^1}{W_k^2}} \right)}\\
{V \_gate = sigmoid \left( {X{W_v^1}{W_v^2}} \right)}\\
{\widetilde K = K \odot K\_gate,{\kern 1pt} {\kern 1pt} {\kern 1pt} {\kern 1pt} {\kern 1pt} {\kern 1pt} {\kern 1pt} {\kern 1pt} \widetilde V = V \odot V\_gate}\\
{O = \left[ {Q\left[ {\widetilde K^T\widetilde V} \right]} \right] \odot G}
\end{array}
\end{equation}

Where $W_k^1 \in {{\mathbb R}^{d_k \times 16}}$, $W_k^2 \in {{\mathbb R}^{16 \times d_k}}$, $W_v^1 \in {{\mathbb R}^{d_v \times 16}}$ and $W_v^2 \in {{\mathbb R}^{16 \times d_v}}$.

We conduct experiments using the Swin-T configuration and the results are shown in Tab.~\ref{AnalysGate}. Compared with SAGA, both alternative gating designs exhibit degraded performance. Among them, the parameterized variant performs better than its non-parameterized counterpart, suggesting that learnable gating parameters can more effectively capture token-level dependencies. Nevertheless, both variants still surpass PolaFormer, validating that the incorporation of a gating mechanism enhances the representational capacity of the $KV$ feature map and improves overall model performance.

\begin{figure}[!t]
\centering
\includegraphics[width=\textwidth]{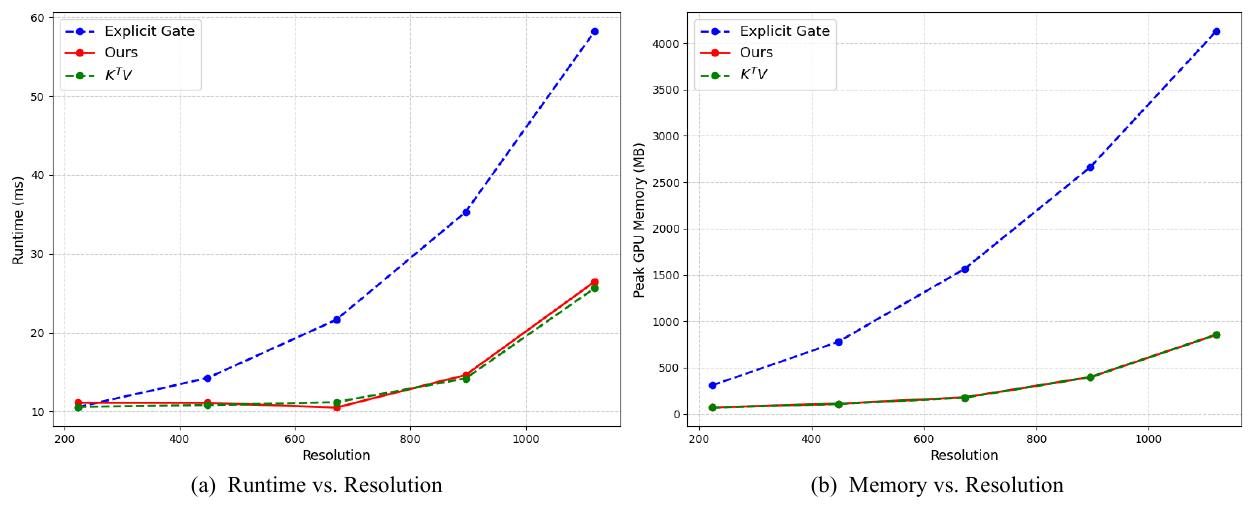}
\caption{The impact of different gate implementations on model efficiency. We measured the runtime and Peak GPU Memory of various gate implementations as a function of resolution on the SAGA-T model, with all data collected on a single NVIDIA 4090 GPU. }
\label{gate_effi}
\end{figure}

\begin{figure}[!t]
\centering
\includegraphics[width=0.6\textwidth]{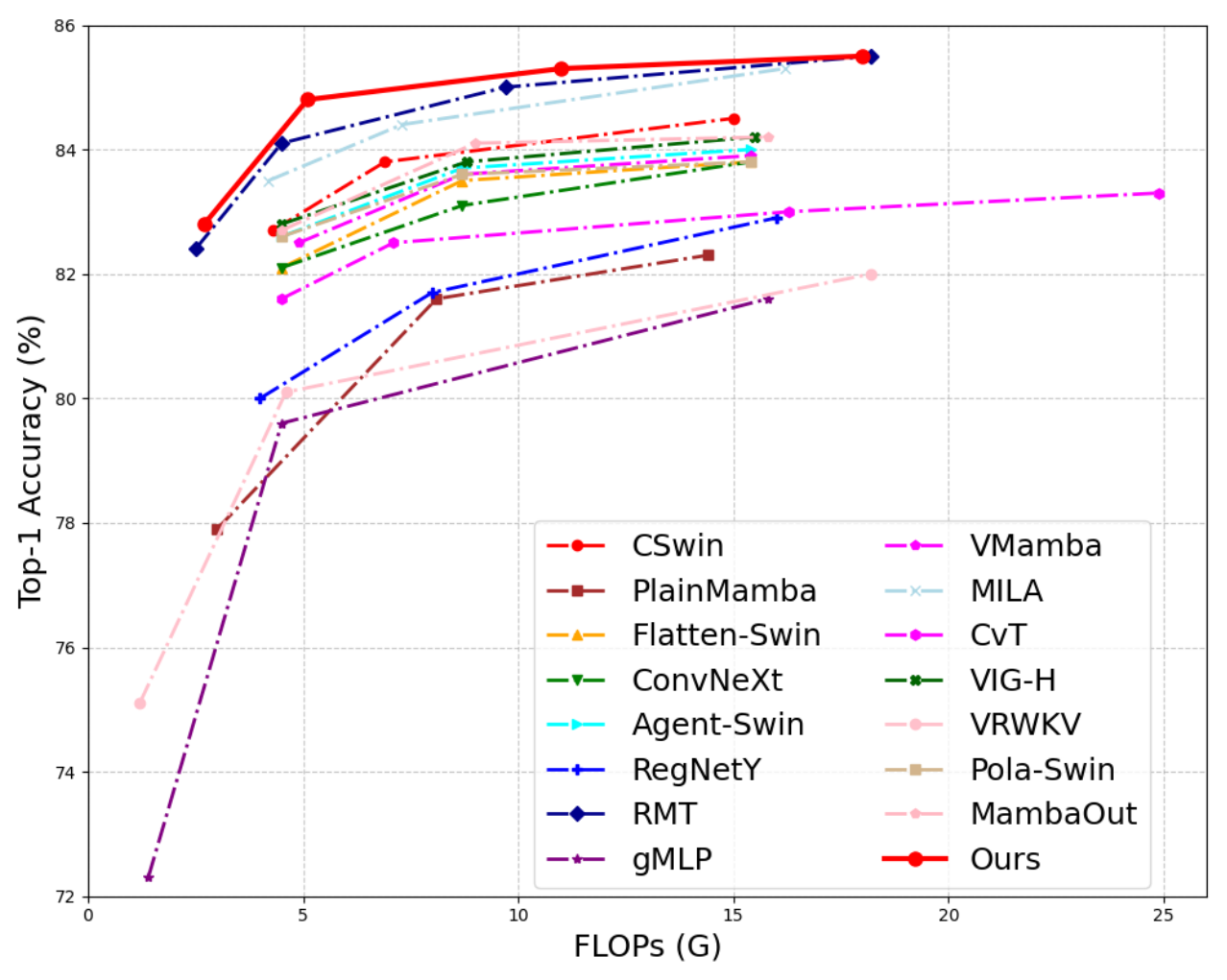}
\caption{Accuracy vs. FLOPs curves on the ImageNet-1K}
\label{Acc Flops}
\end{figure}

\noindent \textbf{Efficiency of Gate. }To further evaluate the computational efficiency of the proposed gating mechanism, we conduct a quantitative analysis on SAGA-T by measuring the runtime and peak GPU memory consumption across different gating implementations. As illustrated in Fig.~\ref{gate_effi}, the \textit{Explicit Gate} variant constructs a unique gating matrix of size \(d_k \times d_v\) for each intermediate SFM, whereas \(K^{T}V\) serves as the baseline method without any gating operation. The results indicate that our proposed Hadamard-product decomposition method achieves nearly the same runtime and memory footprint as the ungated \(K^{T}V\) formulation, while the \textit{Explicit Gate} approach introduces substantial overhead in both runtime and memory usage. These results confirm that the proposed method attains a favorable trade-off between efficiency and performance improvement, validating the practicality of our gating design for large-scale vision models.

\noindent \textbf{Efficiency Analysis. }
Fig.~\ref{Acc Flops} illustrates the Accuracy-FLOPs curves of SAGA compared to other visual backbones. It can be observed that SAGA achieves relatively low FLOPs while maintaining competitive accuracy, thereby striking an excellent balance between efficiency and performance. Furthermore, as evidenced by Fig.~\ref{LLFormer runtime and peak gpu memory}, SAGA maintains lower runtime and peak GPU memory consumption when processing large numbers of tokens, making it particularly suitable for resource-constrained application scenarios.

\begin{figure}[!t]
\centering
\includegraphics[width=\textwidth]{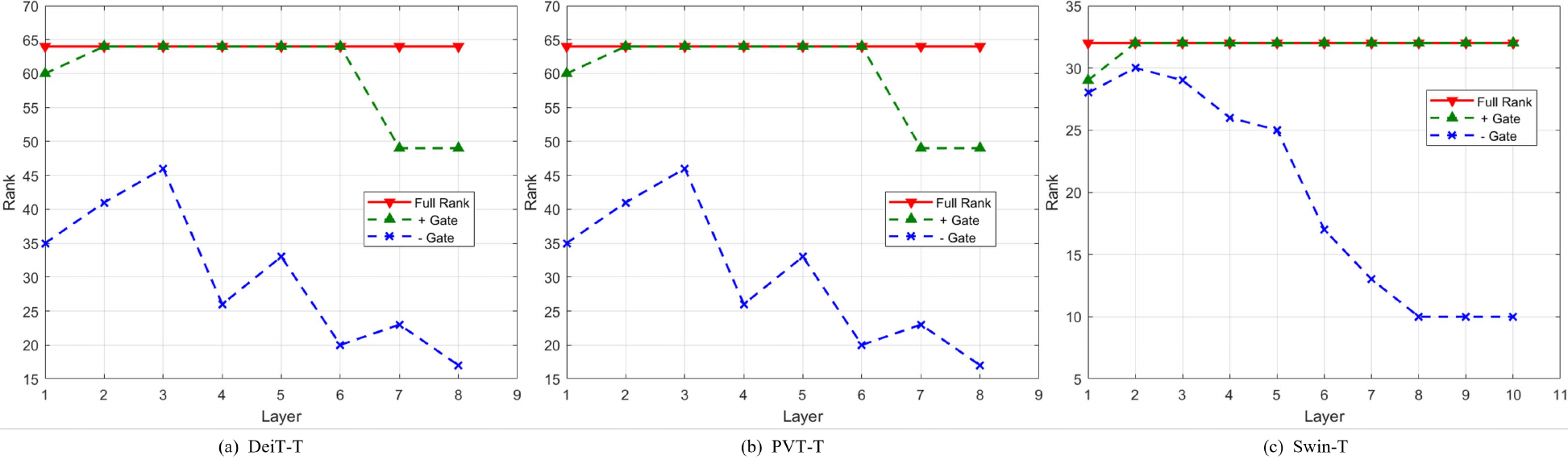}
\caption{The rank changes of $KV$ feature maps for DeiT-T, PVT-T and Swin-T before and after adding the KVGate module.}
\label{Rank Analysis}
\end{figure}

\noindent \textbf{Rank Analysis. }To further verify that the proposed KVGate module increases the rank of the $KV$ feature map and thereby enriches the feature diversity of the semantic repository, we analyze the rank variations of $KV$ feature maps in DeiT-T, PVT-T and Swin-T before and after incorporating KVGate. As shown in Fig.~\ref{Rank Analysis}, prior to integrating KVGate, the ranks of the $KV$ feature maps in these models remain far from full rank and even tend to decrease with increasing network depth. In contrast, after incorporating KVGate, the $KV$ feature maps approach full rank across all layers, indicating a substantial improvement in feature diversity and enhanced expressive capacity of the global semantic repository.

\noindent \textbf{Comparison with Other Linear Attention. }To demonstrate the advantages of our proposed method over other linear attention modules, we conducted comparative experiments using Swin Transformer as baselines. We compared our proposed SAGA with several previous linear attention designs, including efficient attention~\cite{shen2021efficient}, hydra attention~\cite{qin2022cosformer}, enhanced attention~\cite{cai2022efficientvit}, Flatten attention~\cite{han2023flatten} and Polaformer~\cite{meng2025polaformer}.

\begin{table}[!t]
  \centering
    \scalebox{0.75}
    {\begin{tabular}{c|cccc}
    \toprule
    Method & Reso  & Para & FLOPs & Acc(\%)  \\
    \midrule
    Swin-T & $224^2$  & 28M   & 4.4G & 81.2  \\
    HydraAttn & $224^2$  & 29M   & 4.5G & 80.7  \\
    EfficientAttn & $224^2$  & 29M   & 4.5G  & 81.0  \\
    FLattenAttn & $224^2$  & 29M   & 4.5G & 82.1  \\
    PolaFormer & $224^2$  & 29M   & 4.5G & 82.6  \\
    \rowcolor{gray!30} Swin-T-SAGA & $224^2$  &28M   & 4.6G & \textbf{83.0}  \\
    \bottomrule
    \end{tabular}}
    \caption{Comparison of various linear attention methods with Swin-T on the ImageNet-1K dataset.}
  \label{Comparison other linear}%
\end{table}%

As shown in Tab.~\ref{Comparison other linear}, our proposed SAGA significantly outperforms the baselines and other linear attention designs. This indicates that the enhancement of the $KV$ feature map can substantially improve the expressive power of linear attention, thereby achieving superior performance with linear complexity.

\section{Conclusion}
In this work, we present SAGA, a novel and efficient linear attention framework that incorporates an adaptive gating mechanism. By analyzing linear attention from the perspective of the semantic repository, we attribute the performance degradation of existing methods to the low-rank nature of the $KV$ feature map, which limits its ability to capture fine-grained distinctions and relationships among key–value pairs, leading to blurred token representations. To address this issue, we introduce a gating mechanism that adaptively controls the information flow from the intermediate SFMs to the $KV$ feature map, selectively emphasizing relevant features while suppressing irrelevant ones. Furthermore, we propose a Hadamard-product decomposition to efficiently generate the gating matrices, achieving a balanced trade-off between model efficiency and representational power. Through extensive experiments on multiple vision tasks, SAGA demonstrates a favorable balance between efficiency and performance, offering a promising direction for the design of scalable and efficient Transformer architectures.

% \appendix
% \section{Example Appendix Section}\label{app1}


\section*{Declaration of generative AI and AI-assisted technologies in the 
writing process}
During the preparation of this work the authors used ChatGPT in order to improve readability and language. After using this tool, the authors reviewed and edited the content as needed and take full responsibility for the content of the publication.

\section*{Declaration of competing interest}
The authors declare that they have no known competing financialinterests or personal relationships that could have appeared to influencethe work reported in this paper.

\section*{Acknowledgments}
This research was sponsored by Beijing Natural Science Foundation grant number L244050 and L252018.

\section*{Data availability}
All datasets used in this study are publicly available benchmark datasets.
\bibliographystyle{elsarticle-num} 
\bibliography{elsarticle-template-num-ref}




% %% For numbered reference style

%   \textit{\LaTeX: a document preparation system},
%   Addison Wesley, Massachusetts,
% \end{thebibliography}

\end{document}